\pdfoutput=1
\documentclass[11pt]{article}

\usepackage[final]{acl}
\usepackage{graphicx}

\usepackage{times}
\usepackage{latexsym}

\usepackage{amssymb}

\usepackage[T1]{fontenc}
\usepackage[utf8]{inputenc}

\usepackage{microtype}

\usepackage{inconsolata}

\usepackage[normalem]{ulem}
\usepackage{algorithm}
\usepackage{multirow}
\usepackage{amsmath} 
\usepackage{enumitem}

\usepackage{amsmath}
\usepackage{booktabs}

\usepackage{algorithm}
\usepackage{algpseudocode}
\usepackage[most]{tcolorbox}

\newtcolorbox{promptbox}[1]{
    colback=gray!8,          
    colframe=gray!60!black,  
    colbacktitle=gray!60!black, 
    coltitle=white,          
    fonttitle=\normalsize\rmfamily,
    title={#1},             
    sharp corners,
    rounded corners=all,
    arc=4pt,                 
    boxrule=1pt,             
    left=6pt, right=6pt, top=6pt, bottom=6pt,
    breakable                
}

\title{Decoupling Knowledge and Task Subspaces for Composable Parametric Retrieval Augmented Generation}

\usepackage{authblk}
\author[1]{\textbf{Weihang Su}\thanks{swh22@mails.tsinghua.edu.cn}}
\author[1]{\textbf{Hanwen Zhang}\thanks{Contributed equally}}
\author[1]{\textbf{Qingyao Ai}}
\author[1]{\textbf{Yiqun Liu}}

\affil[1]{Department of Computer Science and Technology, Tsinghua University}

\begin{document}
\maketitle

\begin{abstract}

Parametric Retrieval-Augmented Generation (PRAG) encodes external documents into lightweight parameter modules that can be retrieved and merged at inference time, offering a promising alternative to in-context retrieval augmentation.  Despite its potential, many PRAG implementations train document adapters with task-supervised objectives, which may cause each adapter to encode both document-specific facts and reusable task-solving behavior. This entanglement may make adapter composition less reliable: when multiple adapters are merged at inference time, their overlapping task behaviors can accumulate together with document-specific updates, potentially making the merged adapter less stable and less focused on the intended document knowledge. To examine this issue, we explore Orthogonal Subspace Decomposition (OSD), an adapter-training setup that separates reusable task behavior from document-specific knowledge adapters. Concretely, we first train a Task LoRA to capture reusable task behavior, and then train document LoRAs to encode document-specific knowledge in an orthogonal subspace. This setup provides a controlled way to examine how orthogonalizing task and document LoRA updates affects adapter composition in multi-document PRAG. Experiments across multiple knowledge-intensive tasks and model scales suggest that this orthogonalization strategy can improve compositional robustness in parametric RAG, especially when multiple document adapters are merged.
\end{abstract}

\section{Introduction}

Retrieval-Augmented Generation (RAG) has become a standard paradigm for grounding large language models (LLMs) with external knowledge by retrieving relevant documents at inference time and injecting them into the prompt~\cite{lewis2020retrieval}. Despite its effectiveness, standard in-context RAG faces an inherent bottleneck: retrieved knowledge is only provided transiently in context, whereas LLMs often utilize knowledge most effectively when it is represented within their parameters. This limitation has motivated growing interest in \emph{parametric} forms of knowledge injection, where retrieved documents are encoded into loadable parameter modules rather than appended as raw text~\cite{su2025parametric, tan2025dynamic}.
In adapter-based Parametric Retrieval-Augmented Generation (PRAG), each document can be encoded as a lightweight parameter module that is retrieved and composed at inference time, pointing toward a new form of external parametric memory for LLMs~\cite{su2025parametric}.

However, simply parameterizing documents is insufficient to build an effective external parametric memory system. 
The critical missing property is {composability}. 
In existing PRAG-style methods, each document adapter is trained using document-grounded task supervision (e.g., question-answering or fact-verification examples). 
Consequently, the learned adapter does not exclusively encode factual knowledge; it simultaneously encode the task-specific behavior required to utilize that knowledge. 
This creates an {entangled representation} of both ``what the document says'' and ``how to perform the task.'' 
Such entanglement renders document parameterization highly suboptimal in multi-document settings, where multiple retrieved adapters must be merged. 
Instead of purely aggregating complementary knowledge, the model redundantly accumulates partially overlapping task patterns. This leads to parameter interference and degraded performance as the number of composed adapters increases.

To examine this issue, we study \textbf{Orthogonal Subspace Decomposition (OSD)}, an adapter-training setup that explores whether task-level behavior and document-level knowledge can be separated for more stable parametric memory composition. 
The basic idea is to decouple parameterized external memory into two components: a task component, represented by a shared Task LoRA that captures reusable task behavior, and a knowledge component, represented by document-specific Knowledge LoRAs that encode document-level information. 
During training, each Knowledge LoRA is learned on top of a frozen Task LoRA, with an orthogonality constraint that discourages overlap with the task component. 
At inference time, the model uses the corresponding Task LoRA together with the merged Knowledge LoRAs of the retrieved documents. 
This design allows us to examine whether separating task and document updates can reduce interference when multiple document adapters are composed.

We implement this idea with two variants. 
The soft orthogonal variant adds an orthogonal regularization term during Knowledge LoRA training to penalize overlap with the task component. 
The hard orthogonal variant parameterizes the document adapter within the null space of the Task LoRA, enforcing subspace separation by construction. 
Ideally, the shared Task LoRA captures reusable task behavior, while the document-specific Knowledge LoRAs are encouraged to focus more on document-level information rather than redundant task heuristics. 
This separation is intended to make multi-document composition less affected by repeated task-level updates.

We evaluate this training setup across several knowledge-intensive tasks, including open-domain question answering, fact checking, slot filling, and knowledge-grounded dialogue, using standard PRAG benchmarks and expanded KILT-style multi-task settings.
Across different model scales, our experiments suggest that orthogonalizing task and document LoRA updates can improve the stability of parametric RAG when multiple document adapters are merged.
The effect varies across datasets and models, but the decoupled variants are often less sensitive to increasing retrieval depth than entangled parametric baselines.
Additional representation analysis further suggests that orthogonalization changes the geometry of document adapters and, in the soft variant, makes relevant document pairs more distinguishable in parameter space.
Our primary contributions are summarized as follows:
\begin{itemize}[leftmargin=*]
    \item We analyze a potential source of instability in external parametric memory: document adapters trained with task supervision may entangle document-specific knowledge with reusable task behavior, which can affect multi-document adapter composition.

    \item We explore \textbf{Orthogonal Subspace Decomposition} as a simple decoupling strategy that separates a shared Task LoRA from document-specific Knowledge LoRAs. We consider both {soft} and {hard} orthogonal variants to study how different degrees of subspace separation affect adapter composition.
    
    \item We report experiments across multiple knowledge-intensive tasks and model scales, suggesting that separating task and document LoRA updates can improve the robustness of parametric RAG under adapter composition.
\end{itemize}

\section{Problem Formulation}

Let $q$ denote an input query, and let $\mathcal{R}(q)$ return the top-$k$ relevant documents from an external corpus:
\begin{equation}
\mathcal{R}(q) = \{D_1, D_2, \dots, D_k\}.
\end{equation}

The goal is to generate an output $y$ that is grounded in the retrieved evidence, i.e.,
\begin{equation}
P(y \mid q, \mathcal{R}(q)).
\end{equation}

Classic RAG approaches approximate this objective by concatenating retrieved documents into the input context. Parametric RAG instead encodes each document into a document-specific adapter and composes multiple adapters at inference time. While this avoids long-context injection, existing document-wise parametric memory remains difficult to compose: each document adapter is trained from task supervision grounded in that document, and therefore tends to entangle document-specific knowledge with task-specific behavior. As a result, when multiple document adapters are merged, redundant or conflicting task patterns accumulate alongside the desired factual knowledge, causing severe parameter interference. Our objective is to build a {composable} external parametric memory system. Concretely, given a query $q$ and a task type $t$, we aim to learn a shared task module that captures task-general behavior, together with a set of document modules that encode document-specific factual knowledge, such that the retrieved document modules can be stably composed with the task module at inference time.

\section{Methodology}

We study a simple decomposition strategy for parametric retrieval-augmented generation (PRAG), where task-level behavior and document-specific knowledge are represented by separate LoRA modules. 
Rather than treating this decomposition as a guaranteed disentanglement mechanism, our goal is to examine whether separating these two components during training can make document adapters easier to compose at inference time.
Concretely, we first train a shared {Task LoRA} for each task type, and then train document-specific {Knowledge LoRAs} while keeping the corresponding Task LoRA fixed. 
We instantiate this idea with two variants: a soft orthogonality regularizer and a hard orthogonal parameterization.

\subsection{Overview}

For a task type $t$, we first train a Task LoRA, denoted as $\Delta \theta_T^{(t)}$, using task-level supervision that is not tied to any single document. 
This adapter is intended to capture reusable task behavior, such as output format, task-specific reasoning patterns, and general response style. 
Afterwards, for each document $D_i$, we train a document-specific Knowledge LoRA, denoted as $\Delta \theta_{K,i}$, using examples derived from $D_i$. 
During this stage, the Task LoRA $\Delta \theta_T^{(t)}$ is loaded but frozen, so that the Knowledge LoRA is optimized on top of an already available task adapter.

At inference time, given a query $q$ and task type $t$, the retriever returns a set of relevant documents $\{D_1,\ldots,D_k\}$. 
We activate the corresponding Task LoRA once and aggregate the Knowledge LoRAs associated with the retrieved documents:
\begin{equation}
\theta'(q,t)
=
\theta_0
+
\Delta \theta_T^{(t)}
+
\mathrm{Agg}\left(\Delta \theta_{K,1},\ldots,\Delta \theta_{K,k}\right),
\end{equation}
where $\theta_0$ denotes the base model parameters. 
For each adapted layer, the aggregation can be written as
\begin{equation}
\Delta W_K
=
\sum_{i=1}^{k}\alpha_i \Delta W_{K,i},
\qquad
\sum_{i=1}^{k}\alpha_i = 1,
\end{equation}
where $\alpha_i$ may be uniform or determined by retrieval scores. 
This formulation differs from directly merging multiple task-supervised document adapters, since the shared task component is applied only once.

\subsection{Task and Knowledge Modules}

We use the standard Low-Rank Adaptation (LoRA) parameterization for each adapted weight matrix $W \in \mathbb{R}^{d_{\mathrm{out}}\times d_{\mathrm{in}}}$:
\begin{equation}
\Delta W = BA,
\end{equation}
where $A \in \mathbb{R}^{r \times d_{\mathrm{in}}}$, $B \in \mathbb{R}^{d_{\mathrm{out}} \times r}$, and $r$ is the LoRA rank. 
We omit the LoRA scaling factor for notational simplicity.

For task type $t$, the Task LoRA is trained on a task-level corpus $\mathcal{D}_{\mathrm{task}}^{(t)}$:
\begin{equation}
\mathcal{L}_{\mathrm{task}}
=
-\sum_{(x,y)\in\mathcal{D}_{\mathrm{task}}^{(t)}}
\log P_{\theta_0+\Delta\theta_T^{(t)}}(y\mid x).
\end{equation}
The task-level examples are constructed to emphasize general task execution rather than memorization of a particular document. 
This adapter is then kept fixed when training document-specific Knowledge LoRAs.

For each document $D_i$, we construct a document-level training set $\mathcal{D}_{K,i}$ and optimize the corresponding Knowledge LoRA on top of the frozen Task LoRA:
\begin{equation}
\mathcal{L}_{\mathrm{ce}}^{(i)}
=
-\sum_{(x,y)\in\mathcal{D}_{K,i}}
\log P_{\theta_0+\Delta\theta_T^{(t)}+\Delta\theta_{K,i}}(y\mid x).
\end{equation}
This setup does not by itself guarantee perfect separation between task behavior and document knowledge. 
However, it provides a controlled way to study whether document adapters trained in the presence of a fixed task adapter become more suitable for multi-document composition.

\subsection{Orthogonal Knowledge Learning}

A document-specific adapter trained with task supervision may still learn both document information and task-level shortcuts. 
To reduce this overlap, we explore orthogonality constraints between the Task LoRA and each Knowledge LoRA. 
We apply the constraint to the LoRA down-projection matrices $A$, since their row spaces determine the input directions used by the low-rank update.

For a specific adapted layer, let
\begin{equation}
A_T \in \mathbb{R}^{r_T \times d_{\mathrm{in}}},
\qquad
A_{K,i} \in \mathbb{R}^{r_K \times d_{\mathrm{in}}}
\end{equation}
denote the down-projection matrices of the Task LoRA and the $i$-th Knowledge LoRA, respectively. 
The overlap between their row spaces can be measured by the Frobenius norm of their cross inner-product matrix:
\begin{equation}
\left\| A_T A_{K,i}^{\top} \right\|_F^2.
\end{equation}
This quantity is zero when every row direction of $A_{K,i}$ is orthogonal to every row direction of $A_T$.

\paragraph{Soft Orthogonality.}

The first variant adds an orthogonality penalty when training the Knowledge LoRA:
\begin{equation}
\mathcal{L}_{\mathrm{ortho}}^{(i)}
=
\sum_{\ell}
\left\|
A_T^{(\ell)}
A_{K,i}^{(\ell)\top}
\right\|_F^2,
\end{equation}
where $\ell$ indexes adapted layers. 
The full objective for the document module is
\begin{equation}
\mathcal{L}_{\mathrm{know}}^{(i)}
=
\mathcal{L}_{\mathrm{ce}}^{(i)}
+
\lambda
\mathcal{L}_{\mathrm{ortho}}^{(i)},
\end{equation}
where $\lambda$ controls the strength of the orthogonality regularization.

Equivalently, for each layer, the penalty can be computed as
\begin{equation}
\left\|
A_T
A_{K,i}^{\top}
\right\|_F^2
=
\mathrm{tr}
\left(
A_T
A_{K,i}^{\top}
A_{K,i}
A_T^{\top}
\right).
\end{equation}
This soft variant does not strictly prevent overlap, but it discourages the Knowledge LoRA from using the same projection directions as the Task LoRA.

\paragraph{Hard Orthogonality.}

The second variant enforces orthogonality by reparameterizing the Knowledge LoRA projection matrix. 
For each adapted layer, we compute the singular value decomposition of the trained task projection matrix:
\begin{equation}
A_T = U\Sigma V^{\top}.
\end{equation}
Let $\rho_T$ denote the numerical rank of $A_T$, determined by a singular-value threshold. 
We write $V_{\parallel}\in\mathbb{R}^{d_{\mathrm{in}}\times \rho_T}$ for the right singular vectors spanning the row space of $A_T$, and $V_{\perp}\in\mathbb{R}^{d_{\mathrm{in}}\times (d_{\mathrm{in}}-\rho_T)}$ for an orthonormal basis of its null space. 
We then parameterize the Knowledge LoRA down-projection matrix as
\begin{equation}
A_{K,i}
=
\widehat{A}_{K,i} V_{\perp}^{\top},
\end{equation}
where
\begin{equation}
\widehat{A}_{K,i}
\in
\mathbb{R}^{r_K \times (d_{\mathrm{in}}-\rho_T)}
\end{equation}
is the learnable parameter matrix. 
Since $V_{\perp}^{\top}V_{\parallel}=0$, this construction gives
\begin{equation}
A_{K,i}A_T^{\top}=0.
\end{equation}
Thus, the Knowledge LoRA is constrained to use input directions orthogonal to the row space of the Task LoRA.

Compared with the soft variant, this hard variant imposes a stricter structural constraint and removes the need for an auxiliary orthogonality loss. 
However, it also reduces the available parameter space for the Knowledge LoRA. 
We therefore treat the two variants as complementary ways to explore whether reducing task--knowledge overlap improves the composability of parametric memory.

\section{Experimental Setup}

\subsection{Baselines}

To comprehensively evaluate the effectiveness of our proposed decoupling framework, we compare D-PRAG and D-PRAG-hard with two representative retrieval-augmented paradigms: standard in-context RAG~\citep{lewis2020retrieval} and Parametric RAG (PRAG)~\citep{su2025parametric}. 
% In addition, we include a stronger PRAG variant equipped with a shared Task LoRA to examine whether the performance gains of our method can be attributed merely to introducing task-level parameter sharing. 
Further details of the baseline implementations and experimental settings are provided in Appendix~\ref{sec:appendixB1}.

\subsection{Datasets}

We evaluate our method on five task categories, including four knowledge-intensive tasks from the KILT benchmark~\citep{petroni2021kilt} and one vertical-domain setting in medicine. The knowledge-intensive tasks cover open-domain question answering, fact-checking, slot filling, and knowledge-grounded dialogue.

For open-domain question answering, we use the DPR Wikipedia dump as the retrieval corpus~\citep{karpukhin2020dense}, and evaluate on four datasets: 2WikiMultihopQA~\citep{ho2020constructing}, HotpotQA~\citep{yang2018hotpotqa}, ComplexWebQuestions~\citep{talmor2018web}, and PopQA~\citep{mallen2022not}. For fact-checking, slot filling, and dialogue, we use the knowledge sources provided by KILT~\citep{petroni2021kilt}, and select FEVER~\citep{thorne2018fever} for fact-checking, Zero-Shot RE~\citep{levy2017zero} for slot filling, and Wizard of Wikipedia~\citep{dinan2018wizard} for dialogue. For the vertical-domain setting, we use PubMed Abstracts~\citep{gao2020pile} as the corpus and evaluate on PubMedQA~\citep{wu2025medreason}. More details about the datasets and corpora are provided in Appendix~\ref{sec:appendixB2}.

\begin{figure*}[t]
\centering
    \includegraphics[width=\textwidth]{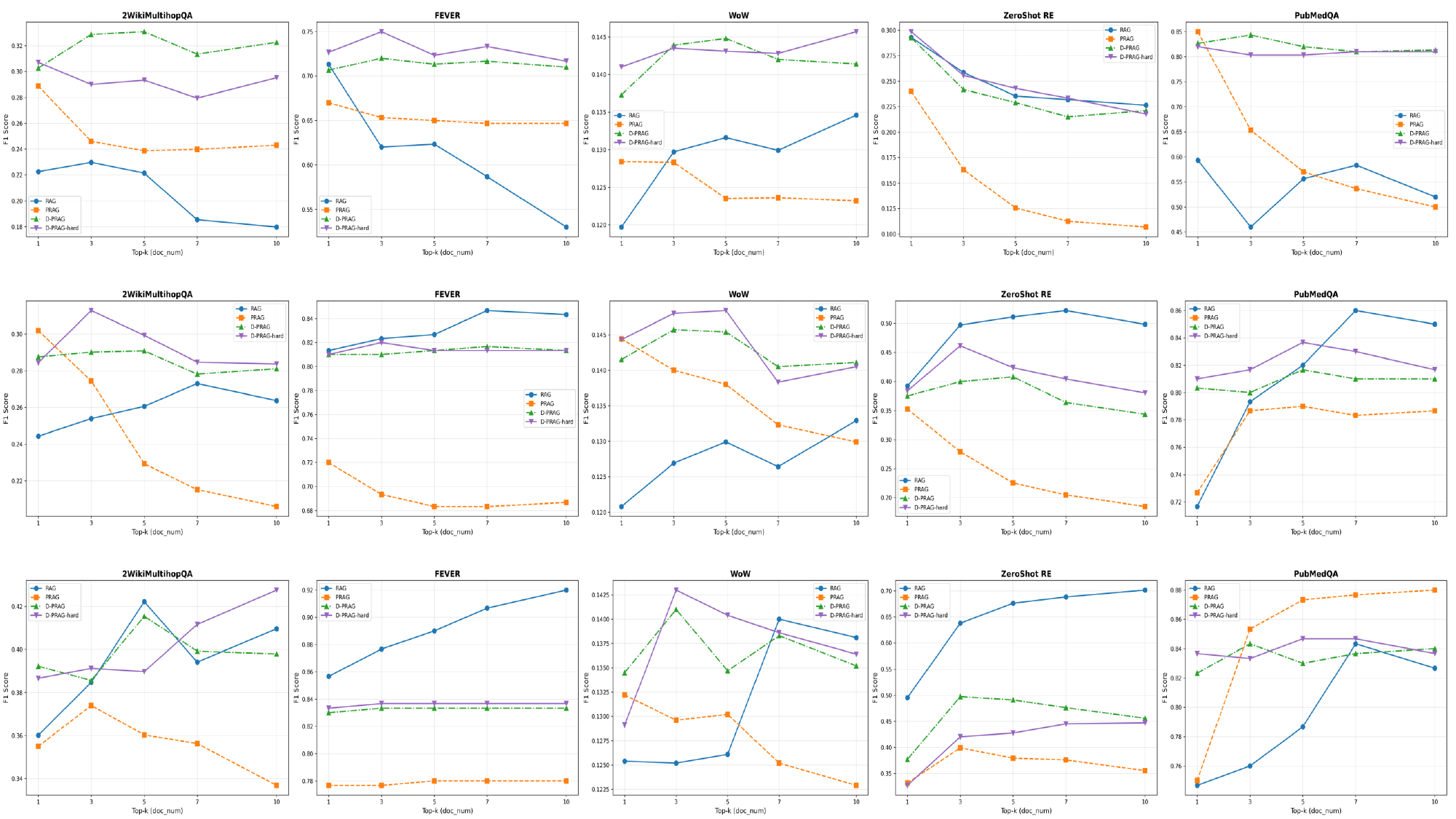}
    \caption{Performance comparison across different retrieval depths ($K \in \{1,3,5,7,10\}$) on five representative datasets.  From top to bottom, the three rows report results on Llama-3.2-1B-Instruct, Llama-3.2-3B-Instruct, and Meta-Llama-3-8B-Instruct. }
    \label{fig:similarity}
\end{figure*}

\subsection{Implementation Details}

We conduct experiments on the Llama 3 model family, including Llama-3.2-1B-Instruct, Llama-3.2-3B-Instruct, and Meta-Llama-3-8B-Instruct. For retrieval, we employ BM25 implemented with ElasticSearch to retrieve the top-$K$ relevant documents for each query. To study the sensitivity of different methods to retrieval depth and context density, we vary the number of retrieved documents with $K \in \{1, 3, 5, 7, 10\}$. All experiments are conducted on NVIDIA A100 (40GB) GPUs. Additional implementation details, including hyperparameter settings, are provided in Appendix~\ref{sec:appendixB3}, and the prompt templates used in our experiments are listed in Appendix~\ref{sec:appendixC}.

\subsection{Evaluation Metrics}

We adopt task-specific evaluation metrics to assess the performance of our decoupling framework. Specifically, we report F1 for open-domain question answering, slot filling, and dialogue, and Accuracy for fact-checking and the medical-domain task. Due to the computational cost of large-scale inference, we evaluate all methods on a representative subset comprising the first 300 test instances from each dataset, thereby providing a consistent comparison setting across models and methods. Further details on the evaluation protocols and metric definitions are provided in Appendix~\ref{sec:appendixB4}.

\begin{figure}[t]
\centering
    \includegraphics[width=\columnwidth{}{}]{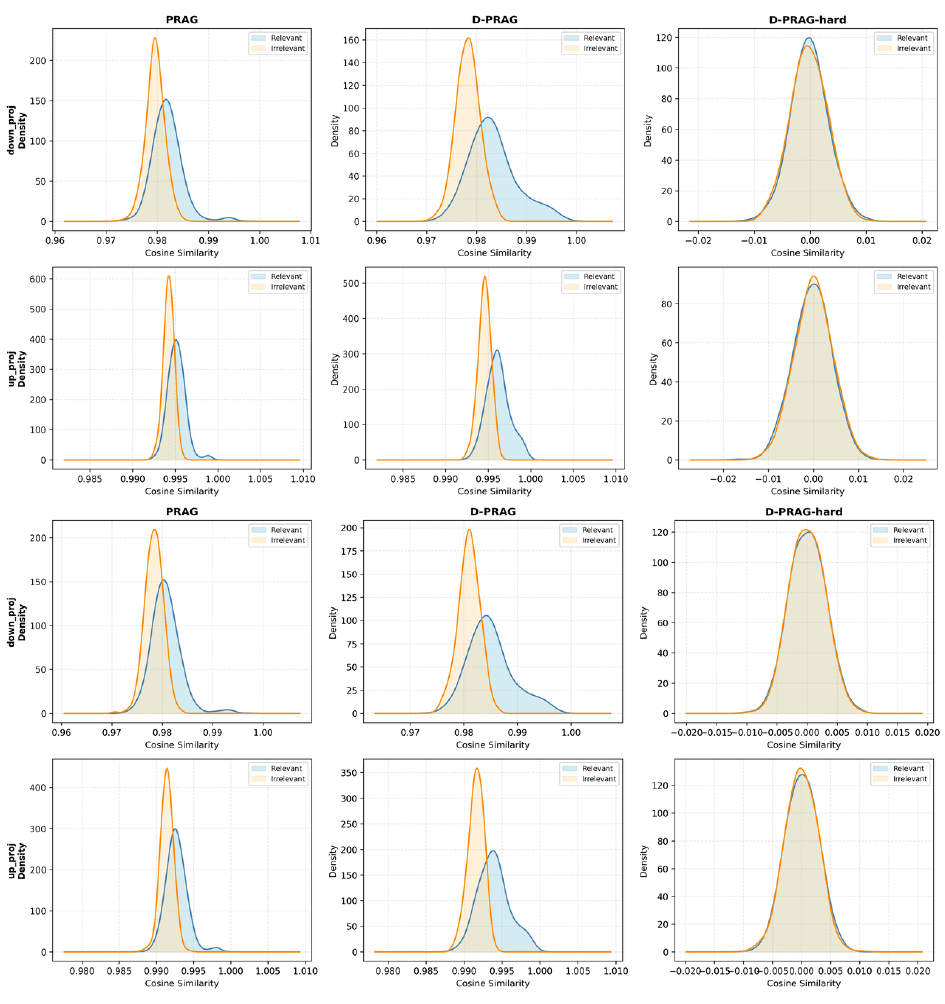}
    \caption{Cosine similarity distributions of relevant and irrelevant passage pairs under PRAG, D-PRAG, and D-PRAG-hard in the same experimental setting as our main results. The upper and lower parts correspond to Llama-3.2-1B-Instruct and Llama-3.2-3B-Instruct, respectively. Similarity is computed from flattened document-specific LoRA parameters, shown separately for the \texttt{down\_proj} and \texttt{up\_proj} matrices.}
    \label{fig}
\end{figure}

\section{Experimental Results}

\subsection{Main Results}

Figure~\ref{fig:similarity} compares standard RAG, PRAG, D-PRAG, and D-PRAG-hard as the number of retrieved documents increases under three similarity-controlled retrieval settings. 
Overall, PRAG is often more sensitive to retrieval depth than the decoupled variants: in several settings, its performance decreases as more document-specific adapters are merged. 
By contrast, D-PRAG and D-PRAG-hard generally exhibit flatter curves, suggesting that separating task-level behavior from document-level information can improve the stability of multi-document adapter composition. 
The effect is not uniform across all datasets and model scales: standard RAG remains competitive in some cases, and the relative performance of D-PRAG and D-PRAG-hard varies across settings. 
Nevertheless, the repeated pattern that the decoupled variants are less affected by increasing $K$ is consistent with our hypothesis that reducing task-document entanglement can make external parametric memory more robust under adapter composition.

\subsection{Representation Analysis}

To further inspect how different training strategies affect the geometry of document adapters, we compare the cosine similarity distributions between relevant and irrelevant passage pairs using flattened document-specific LoRA parameters, as shown in Figure~\ref{fig}. 

For PRAG, both relevant and irrelevant pairs are concentrated in a highly positive similarity range, indicating that document adapters tend to share strong common directions in parameter space. 
This makes the similarity distributions only moderately informative for distinguishing relevance, and is consistent with the possibility that task-supervised document adapters contain shared task-level components in addition to document-specific information.

D-PRAG shows a clearer separation between relevant and irrelevant pairs in this analysis: relevant pairs are generally shifted toward higher cosine similarity, while irrelevant pairs remain more concentrated at lower values. 
This suggests that reducing overlap with the task component can make document adapters somewhat more sensitive to document-level relatedness. 
However, the distributions still overlap, so we interpret this result as suggestive rather than as direct evidence that the learned adapters encode purely factual knowledge.

D-PRAG-hard exhibits a different pattern: the cosine similarities of both relevant and irrelevant pairs are centered near zero and largely overlap. 
This is expected from the hard orthogonal parameterization, which strongly changes the scale and geometry of adapter directions. 
Therefore, raw cosine similarity in the flattened parameter space becomes less directly interpretable as a relevance indicator for this variant. 
Overall, the representation analysis provides an auxiliary view consistent with our main hypothesis: separating task-level and document-level updates can change the geometry of document adapters, and in the soft variant this change is associated with clearer relevance-related structure.

\section{Related Work}

\paragraph{Retrieval-Augmented Generation and Parametric Knowledge Injection.}
Standard RAG~\cite{lewis2020retrieval} and its enhancements~\cite{izacard2021leveraging, izacard2023atlas, borgeaud2022improving} rely on \emph{in-context} knowledge injection, which is constrained by inference context windows. To address this, parametric knowledge injection stores external knowledge directly in model weights. Early works used dedicated adapters~\cite{wang2021kadapter}, while recent Parametric RAG (PRAG) and DyPRAG encode documents into composable or dynamically generated LoRA modules~\cite{su2025parametric, tan2025dynamic}. Our work shares this parametric focus but targets a critical bottleneck: the limited \emph{composability} of document-wise modules caused by the entanglement of task behavior and factual knowledge.

\paragraph{Parameter-Efficient Adaptation and Modular Composition.}
Parameter-efficient fine-tuning (PEFT) methods like adapters and LoRA~\cite{houlsby2019parameter, hu2022lora} naturally enable modular composition. Existing approaches combine these lightweight modules or parameter updates across different tasks or languages~\cite{pfeiffer2021adapterfusion, pfeiffer2020madx, huang2023lorahub, ilharco2023editing, yadav2023ties}. However, these methods primarily focus on \emph{task-} or \emph{language-level} adaptations. In contrast, our work tackles the distinct challenge of composing multiple \emph{document-level} modules for parametric RAG, where modules must contribute complementary factual content without redundant task heuristics.

\paragraph{Orthogonality and Interference-Aware Low-Rank Adaptation.}
Orthogonal subspace design effectively reduces parameter interference in continual learning and optimizes LoRA training~\cite{wang2023orthogonal, buyukakyuz2024olora}. While these works validate orthogonality for improving modularity, they focus on sequential task adaptation and optimization efficiency. We fundamentally repurpose orthogonal constraints for \emph{functional} decomposition in external parametric memory: separating a shared task subspace (for reusable execution behavior) from orthogonal document-specific subspaces (for factual memory) to ensure stable multi-document merging.

\section{Conclusion}

In this paper, we studied whether external parametric memory can be made more composable by separating task-level behavior from document-level information. 
Motivated by the observation that document adapters trained with task supervision may encode both document-specific information and reusable task behavior, we explored \textbf{Orthogonal Subspace Decomposition}, an adapter-training setup that combines a shared Task LoRA with document-specific Knowledge LoRAs. 
We instantiated this idea with soft and hard orthogonal variants to examine how different forms of task-document separation affect multi-document adapter composition.
Across multiple knowledge-intensive tasks, datasets, and model scales, our experiments suggest that orthogonalizing task and document LoRA updates can improve the stability of parametric RAG when multiple document adapters are merged. 
The effect varies across settings, and the decoupled variants are not uniformly superior, but they often show lower sensitivity to retrieval depth than entangled parametric baselines. 
Additional representation analysis suggests that orthogonalization changes the geometry of document adapters and, in the soft variant, can make relevant document pairs more distinguishable in parameter space. 
These findings position task-document decoupling as a useful direction for further studying composable external parametric memory.

\section{Limitations}

This work should be understood as a preliminary technical report rather than a finalized methodological contribution. 
Our goal is to document an exploratory empirical study of task-knowledge LoRA decomposition for parametric retrieval-augmented generation, focusing on whether separating task-level behavior from document-specific knowledge can improve adapter composability in initial experiments. 
Although the results suggest that such a decomposition may help stabilize multi-document adapter composition, they should be interpreted as initial empirical observations rather than definitive evidence of a fully established solution. 
Further studies are needed to validate the robustness of this strategy across broader model families, larger-scale retrieval settings, and alternative adapter architectures.

% \clearpage 
\bibliography{custom}
\appendix

\section{Experimental setup}
\label{sec:appendixB}

\subsection{Baselines}
\label{sec:appendixB1}
\begin{itemize}[leftmargin=*]
    \item \textbf{Standard RAG}\citep{lewis2020retrieval}: This is a traditional in-context retrieval-augmented generation method. It directly prepends the top-$K$ retrieved documents to the input prompt, enabling the model to leverage external knowledge through in-context learning without any modification to the model parameters.
    \item \textbf{Parametric RAG}\citep{su2025parametric}: Unlike the standard RAG, this approach internalizes external knowledge by parameterizing retrieved documents into LoRA\cite{hu2022lora} modules. It enables parametric knowledge injection, allowing the model to memorize the retrieved content within its weights during the inference stage. We utilize the official implementation from the PRAG library.
    % \item \textbf{PRAG equipped with task\_LoRA}: To ensure a fair comparison, we extend the basic Parametric RAG by incorporating a task-specific LoRA adapter. This baseline represents a straightforward combination of knowledge parameterization and task adaptation. However, unlike our proposed D-PRAG, it lacks an explicit decoupling mechanism, meaning the knowledge and task-specific updates may still suffer from mutual parameter interference. 
\end{itemize}

\subsection{Datasets and Corpus}
\label{sec:appendixB2}
\textbf{Open domain QA}:We use the Wikipedia dump provided by DPR\citep{karpukhin2020dense} as our corpus.To evaluate the model's capabilities in both knowledge retrieval and complex reasoning, we select four diverse and representative benchmarks:
\begin{itemize}[leftmargin=*]
\item \textbf{2WikiMultihopQA(2WQA)}\citep{ho2020constructing}: A Wikipedia-based multi-hop QA dataset that emphasizes the ability to associate and reason over cross-document factual relationships.
\item \textbf{HotpotQA(HQA)}\citep{yang2018hotpotqa}: A multi-hop reasoning benchmark that requires models to synthesize evidence across multiple documents to derive the correct answer.
\item \textbf{ComplexWebquestions(CWQ)}\citep{talmor2018web}: Comprising complex natural language queries from the web, this dataset demands compositional reasoning over multiple sub-questions.
\item \textbf{PopQA}\cite{mallen2022not}:An open-domain QA dataset focused on popular entities, where questions typically rely on single-hop retrieval or implicit factual knowledge.
\end{itemize}
\textbf{Fact-Checking}: We use the knowledge source provided by KILT\cite{petroni2021kilt} as our external corpus.
\begin{itemize}[leftmargin=*]
    \item \textbf{FEVER} \citep{thorne2018fever}: A large-scale benchmark for fact extraction and verification, requiring the model to classify claims as Supported, Refuted, or NotEnoughInfo based on retrieved evidence.
\end{itemize}
\textbf{Slot-filling}: We use the knowledge source provided by KILT\cite{petroni2021kilt} as our external corpus.
\begin{itemize}
    \item \textbf{Zero Shot RE} \citep{levy2017zero}: A relation extraction dataset formulated as a slot-filling task, which evaluates the model's ability to extract specific subject-relation-object triplets from the corpus.
\end{itemize}
\textbf{Dialogue}: We use the knowledge source provided by KILT\cite{petroni2021kilt} as our external corpus.
\begin{itemize}
    \item \textbf{Wizard of Wikipedia (WoW)} \citep{dinan2018wizard}: A knowledge-driven dialogue dataset where the model must engage in conversations by retrieving and incorporating relevant Wikipedia topics into its responses.
\end{itemize}
\textbf{Medical-Verify}: To test the effectiveness of our proposed method in vertical domains, we conduct experiments in Medicine. We use the PubMed Abstracts subset of The Pile\cite{gao2020pile} as our external corpus.  
\begin{itemize}
    \item \textbf{PubMedQA} \citep{wu2025medreason}: A specialized biomedical QA dataset based on PubMed abstracts. It requires models to answer research questions by reasoning over the provided context. 
\end{itemize}

\subsection{Implementation Details}
\label{sec:appendixB3}
\begin{itemize}[leftmargin=*]
    \item \textbf{Data Augmentation}: During Document Augmentation Process, We select meta-Llama-3-8B-Instruct. For knowledge-intensive tasks, we employ a multi-task data augmentation strategy. For each document, we generate three input-output pairs for each of the four task types—Open domain QA, Fact-checking, Slot-filling, and Dialogue—specifically for training the Doc\_LoRA. To further train the task\_LoRA, we generate an additional single pair per task type from the same document. In the vertical domain (Medical), the augmentation is focused on domain-specific verification. Each document yields three Medical Verification pairs for Doc\_LoRA training and one pair for the task\_LoRA. The prompt templates we use in data augmentation are provided in Appendix\ref{sec:appendixC1}
    \item \textbf{Task LoRA Training}: For knowledge-intensive tasks, we randomly sample 1,500 documents from the augmented corpus, while for vertical domains, 900 documents are sampled to construct the training sets. The complete training data are available at our official repository: \footnote{\url{https://github.com/oneal2000/OSD}},
During the training process, we adopt a default configuration of 1 epoch and a learning rate of $1 \times 10^{-4}$. Specific adjustments are made for optimal convergence on certain tasks: 
(1) For the \textbf{1B model}, we set the epoch to 3 for Fact-checking and 2 for Dialogue; 
(2) For the \textbf{3B model}, we utilize a learning rate of $8 \times 10^{-5}$ for Fact-checking and set the epoch to 2 for Dialogue. 
% The prompt templates we use to train task LoRA are provided in Appendix\ref{sec:appendixC2}

    \item \textbf{Doc LoRA Training}: In the knowledge parameterization stage, we set the default configuration to 1 epoch with a learning rate of $3 \times 10^{-4}$. To ensure robust performance across different tasks and model scales, we apply the following specific adjustments:(1) For PopQA, WoW, and Zero Shot RE, the training epoch is increased to 2 for all model scales (1B, 3B, and 8B).(2) On the 3B model, the learning rate is adjusted to $1 \times 10^{-4}$ for FEVER and $5 \times 10^{-4}$ for Zero Shot RE. On the 8B model, the learning rate for FEVER is set to $5 \times 10^{-5}$, while Zero Shot RE utilizes $5 \times 10^{-4}$.

    For a fair comparison, the training hyperparameters remain identical across PRAG, D-PRAG, and D-PRAG-hard under the same experimental settings.For our decoupling mechanisms, in the soft orthogonality approach (D-PRAG), we set the regularization coefficient $\lambda$ to 0.1 for 1B and 3B models, and 0.2 for the 8B model. In the hard orthogonality approach (D-PRAG-hard), the threshold $\tau$ to identify near-zero singular values in the Singular Value Decomposition (SVD) is set to $1 \times 10^{-5}$. The prompt templates we use to train Doc LoRA are provided in Appendix\ref{sec:appendixC}

    \item \textbf{Inference}: To ensure a fair and reproducible evaluation, the decoding temperature is strictly set to 0 (i.e., greedy search) for all generation tasks. The prompt templates we use during the inference process are provided in Appendix\ref{sec:appendixC}
\end{itemize}

\subsection{Evaluation and Metrics}
\label{sec:appendixB4}
To ensure a comprehensive and computationally efficient evaluation, we report the performance on the first 300 instances of the overall test split for each dataset. Furthermore, for the multi-hop reasoning benchmarks (2WQA and HQA), we conduct an additional  evaluation by testing the first 300 instances within each specific sub-category. We adopt task-specific metrics for this evaluation: F1-score is reported for the QA (2WQA, HQA, CWQ, PopQA), Slot-filling (Zero Shot RE), and Dialogue (WoW) tasks, while Accuracy is utilized for Fact-checking (FEVER) and Medical Verification (PubMedQA).

\section{Prompt Templates}
\label{sec:appendixC}

In this section, we detail the prompt templates utilized across our framework, divided into two main stages: Data Augmentation and Training/Inference.

\subsection{Data Augmentation Prompts}
During the data augmentation phase, the model is prompted to generate a rewritten version of the provided text and structured input-output pairs. 
\textbf{Rewrite Prompt}: This prompt template is used to rewrite the provided text.

\begin{promptbox}{Rewrite Prompt}
Rewrite the following passage. While keeping the entities, proper nouns, and key details such as names, locations, and terminology intact, create a new version of the text that expresses the same ideas in a different way. Make sure the revised passage is distinct from the original one, but preserves the core meaning and relevant information.
{[passage]}
\end{promptbox}
\noindent\textbf{Augmentation Prompts}: The augmentation prompt templates provided below are used to generate four task-specific input-output pairs per passage, with three allocated for training the Doc LoRA and the remaining one reserved for the task LoRA.
\noindent{[}...{]} serves as a placeholder representing a concrete JSON instance of the specific task type. Readers are encouraged to refer to our official GitHub repository (\footnote{\url{https://github.com/oneal2000/OSD}}) for the complete executable examples.
\label{sec:appendixC1}

\begin{promptbox}{Augmentation Prompt: Open-domain QA}
I will provide a passage of text, and you need to generate four different questions based on the content of this passage. Each question should be answerable using the information provided. Additionally, please provide an appropriate answer for each question derived from the passage. \\
You need to generate the question and answer in the following format: \\
{[}...{]} \\
This list should have at least four elements. \\
Passage: \\
{[passage]}
\end{promptbox}

\begin{promptbox}{Augmentation Prompt: Fact-Checking}
I will provide a passage of text, and you need to generate four claims based on the content of this passage. Each claim should be verifiable using the information provided in the passage. Additionally, please provide an appropriate label for each claim, indicating whether it is 'SUPPORTS' or 'REFUTES'. \\
You need to generate each claim and label in the following format: \\
{[}...{]} \\
This list should have at least four elements. \\
Passage: \\
{[passage]}
\end{promptbox}

\begin{promptbox}{Augmentation Prompt: Slot-filling}
I will provide a passage of text, and you need to extract four slot-filling examples from it. \\
Each example should identify a subject entity mentioned in the passage, one of its relations, and the corresponding object entity. \\
You should model the input as a structured string in the format 'subject\_entity [SEP] relation'. \\
The output should be the object entity that fills the slot, based on the passage. \\
Additionally, for each slot-filling example, you need to generate a natural language template question that could be answered by the output (use the subject entity and relation in the question)  \\
You need to generate the input, output, and the template question in the following format:\\
{[}...{]} \\
You only need to output this list in the above format.\\
This list should have at least four elements\\
Passage: \\
{[passage]}
\end{promptbox}

\begin{promptbox}{Augmentation Prompt: Dialogue}
I will provide a passage of text from Wikipedia, and you need to generate four knowledge-grounded dialogues in the style of the Wizard of Wikipedia dataset. \\
Each dialogue should be a natural, multi-turn conversation between a curious user and a knowledgeable assistant (wizard) who has access to the provided passage. \\
The assistant should provide informative, detailed responses based on the passage content, while maintaining a natural conversational flow. \\
The input should contain the conversation history (alternating user and assistant messages), and the output should be the assistant's response to the last user message. \\
Each message in the input should be separated by a newline character (\textbackslash n), except the last message. \\
The assistant's responses should be informative, engaging, and naturally incorporate information from the passage without directly copying it. \\
The user's questions should progressively explore different aspects of the topic, building upon previous turns in the conversation. \\
You need to generate the dialogues in the following format:\\
{[}...{]} \\
Important guidelines:\\
- The assistant should provide detailed, informative responses based on the passage\\
- The conversation should feel natural and engaging, not robotic\\
- Each dialogue should have a varied number of turns (typically 1-6 turns)\\
- The user's questions should explore different aspects of the topic mentioned in the passage\\
- The assistant's responses should synthesize information from the passage in a natural way\\
- Generate at least four dialogues with different conversation flows\\
You only need to output this list in the above format.\\
This list should have at least four elements.\\
Passage: \\
{[passage]}
\end{promptbox}

\begin{promptbox}{Augmentation Prompt: Medical Verification}
I will provide a passage. Based ONLY on the factual content of the passage, generate four yes/no questions. \\
Each question must be objectively answerable as 'yes' or 'no' based on the passage. \\
For each question, provide the correct answer as exactly one lowercase word: 'yes' or 'no'. \\
{[}...{]} \\
This list must contain at least four elements. \\
\textbf{Passage:} \\
{[passage]}
\end{promptbox}

\subsection{Training and Inference Prompts}
To maintain strict alignment between the training and inference stages, we employ identical prompt templates throughout both phases. Specifically, during the training of the task LoRA, external passages are deliberately excluded from the prompts to focus entirely on task format adaptation. Conversely, during the Doc LoRA training, we utilize a mixed prompt strategy, incorporating instances both with and without the contextual passages to ensure robust knowledge internalization.

\noindent Specifically, the \texttt{[passages]} placeholder is dynamically populated with a concatenated text of retrieved documents when external context is required; otherwise, it is simply left blank. The concatenated text format is provided here:
\begin{promptbox}{Concatenated Text}
Passage 1: {Passage 1}\\
Passage 2: {Passage 2}\\
……\\
Passage K: {Passage K}
\end{promptbox}

\noindent\textbf{Prompt for Open domain QA}: This template is used for Open domain QA.

\begin{promptbox}{Prompt 1: Open domain QA}
You should answer the question by referring to the knowledge provided below and integrating your own knowledge. \\
{[}passages{]} \\

\noindent Question: [question] \\
Answer:
\end{promptbox}
\noindent\textbf{Prompt for Fact-checking}: This template is used for Fact-checking.

\begin{promptbox}{Prompt 2: Fact-Checking}
You are tasked with verifying a claim using the knowledge provided below combined with your own knowledge. \\
Your response MUST be exactly one word: 'SUPPORTS' or 'REFUTES'. Do not output anything else and do not explain your choice. \\
Using the passages and your knowledge, output 'SUPPORTS' if the claim is true, or 'REFUTES' if the claim is false. \\
{[}passages{]} \\

\noindent Claim: [input] \\
Output: 
\end{promptbox}
\noindent\textbf{Prompt for Slot-filling}: This template is used for Slot-filling.
\begin{promptbox}{Prompt 3: Slot-Filling}
You are tasked with extracting the object entity that completes a given slot using the knowledge provided below together with your own knowledge. \\
The input format is: 'subject\_entity [SEP] relation'. \\
You must return only the object entity that is directly connected to the subject\_entity through the specified relation. \\
The extracted entity should be the one that can directly serve as the answer to the question: \\
{[}template\_question{]} \\
Do not provide explanations or additional text, only output the object entity. \\
{[}passages{]} \\

\noindent Input: [input] \\
Output: 
\end{promptbox}
\noindent\textbf{Prompt for Dialogue}: This template is used for Dialogue.
\begin{promptbox}{Prompt 4: Dialogue}
You are tasked with generating a response to the user's last message in the whole conversation history, using the knowledge provided below combined with your own knowledge. \\
The input contains the entire conversation history between the user and the assistant. \\
{[}passages{]} \\

\noindent Input: [input] \\
Output:
\end{promptbox}
\noindent\textbf{Prompt for Medical Verification}: This template is used for Medical Verification.
\begin{promptbox}{Prompt 5: Medical Verification}
You are tasked with answering a medical yes/no question using the knowledge provided below combined with your own medical knowledge. \\
Your response MUST be exactly one word: 'yes' or 'no'. Do not output anything else and do not explain your choice. \\
Using the passages and your knowledge, Output 'yes' if it is true, otherwise output 'no'. \\
{[}passages{]} \\

\noindent Question: [question] \\
Answer:
\end{promptbox}
\end{document}